\documentclass[sigconf]{acmart}

\renewcommand\footnotetextcopyrightpermission[1]{} 

\AtBeginDocument{%
  }

\setcopyright{acmlicensed}
\copyrightyear{2026}
\acmYear{2026}
\acmDOI{XXXXXXX.XXXXXXX}
\acmISBN{978-1-4503-XXXX-X/2026/10}

\usepackage{hyperref}
\usepackage{url}
\usepackage{booktabs}
\usepackage{tabularx}
\usepackage{array}
\usepackage{graphicx}
\usepackage[table]{xcolor}
\usepackage{mdframed}
\usepackage{listings}
\usepackage{subcaption}
\usepackage{pifont}
\usepackage{algorithm}
\usepackage{algpseudocode}
\usepackage{multirow}
\usepackage{enumitem}
\usepackage{capt-of}

\usepackage{bm}

\usepackage{cuted}
\usepackage{capt-of}
\usepackage{etoc}
\usepackage{wrapfig}

\usepackage{booktabs}   
\usepackage{xcolor}     
\usepackage{colortbl}   
\usepackage{makecell}   
\usepackage{xspace}
\definecolor{DGFMBlue}{HTML}{4C7FAF}
\definecolor{DGFMCoral}{HTML}{F05252}

\definecolor{DGFMGain}{HTML}{12AFC0}
\definecolor{DGFMLoss}{HTML}{F05252}

\definecolor{DGFMReference}{HTML}{F7F5F2}
\definecolor{DGFMHighlight}{HTML}{DCEBFC}

\definecolor{DGFMBackground}{HTML}{FFFCF4}
\definecolor{DGFMGrid}{HTML}{E8E3D8}

\newcounter{dgfmobservation}

\newcommand{\method}{CL-DGFM\xspace}

\begin{document}

\makeatletter
\def\@authorbaselinestretch{0.9}
\makeatother

\title{A Common Language for Dynamic Graphs: Interaction Tokenization Enables Foundation Models}


\author{Fuyuan Ma}
\authornote{Both authors contributed equally to this research.}
\email{mafy23@mails.jlu.edu.cn}
\affiliation{%
  \institution{School of Artificial Intelligence, Jilin University}
  \city{Changchun}
  \state{Jilin}
  \postcode{130012}
  \country{China}
}

\author{Yili Wang}
\authornotemark[1]
\email{wangyili@jlu.edu.cn}
\affiliation{%
  \institution{School of Artificial Intelligence, Jilin University}
  \city{Changchun}
  \state{Jilin}
  \postcode{130012}
  \country{China}
}

\author{Xin He}
\email{hexin20@mails.jlu.edu.cn}
\affiliation{%
  \institution{School of Artificial Intelligence, Jilin University}
  \city{Changchun}
  \state{Jilin}
  \postcode{130012}
  \country{China}
}

\author{Xin Zhang}
\email{xzhang25@mails.jlu.edu.cn}
\affiliation{%
  \institution{School of Artificial Intelligence, Jilin University}
  \city{Changchun}
  \state{Jilin}
  \postcode{130012}
  \country{China}
}

\author{Ying Wang}
\authornote{Corresponding author.}
\email{wangying2010@jlu.edu.cn}
\affiliation{%
  \institution{School of Computer Science and Technology, Jilin University}
  \city{Changchun}
  \state{Jilin}
  \postcode{130012}
  \country{China}
}

\author{Xin Wang}
\email{xinwang@jlu.edu.cn}
\affiliation{%
  \institution{School of Artificial Intelligence, Jilin University}
  \city{Changchun}
  \state{Jilin}
  \postcode{130012}
  \country{China}
}
\renewcommand{\shortauthors}{Fuyuan Ma and Yili Wang, et al.}

\begin{abstract}
Dynamic graphs capture evolving interactions in Web systems, including online communication, user engagement, and recommendation. 
Graph foundation models aim to reuse predictive capabilities across networks, but extending this capability to dynamic graphs faces two challenges. First, node identities are graph-specific, and similar interaction patterns can appear at different temporal and activity scales, making histories difficult to represent consistently across networks. Second, the predictive roles of historical patterns vary across interaction dynamics, limiting transfer from a narrow range of training processes. In this paper, we show that a common interaction language, combined with learning across diverse interaction processes, offers a basis for dynamic graph foundation modeling. We propose \method{}, a Common Language Dynamic Graph Foundation Model for temporal link prediction. Its Dynamic Interaction Tokenizer encodes interaction memory, historical connectivity, and ordered destination transitions into calibrated tabular inputs, making dynamic histories accessible to existing predictors without an additional trainable graph encoder. Its Dynamic Interaction Prior trains a pretrained tabular foundation predictor across diverse synthetic dynamics. Labeled historical context guides prediction, with a Contextual Value Adapter modulating attention values. The model transfers to real graphs without target-graph parameter updates. Experiments on TGB and TGB-Seq demonstrate competitive performance. Further comparisons show that the common language supports basic machine learning and neural models, and enables direct use of frozen tabular foundation models on dynamic graphs.
\end{abstract}

\vspace{-1.5em}

\vspace{-1.5em}


\received{20 February 2007}
\received[revised]{12 March 2009}
\received[accepted]{5 June 2009}

\maketitle
\vspace{-1.5em}

\section{Introduction}
\label{sec:introduction}

Dynamic graphs capture the evolving interactions underlying Web systems, including online communication, user engagement, and recommendations. Temporal link prediction uses past interactions to predict which entities will interact at a given time. Existing temporal graph models are typically trained separately for each dataset, requiring new training to serve networks with different entities and activity patterns. Yet these networks can share recurring interaction patterns, including repeated contacts, indirect connections, and ordered destination visits. Such regularities raise the possibility of learning predictive knowledge that can be reused across networks. This leads to a fundamental question: \textbf{can we build a dynamic graph foundation model (DGFM) that predicts across networks without dataset-specific retraining?}

\begin{figure}[t]
\centering
\includegraphics[width=\linewidth]{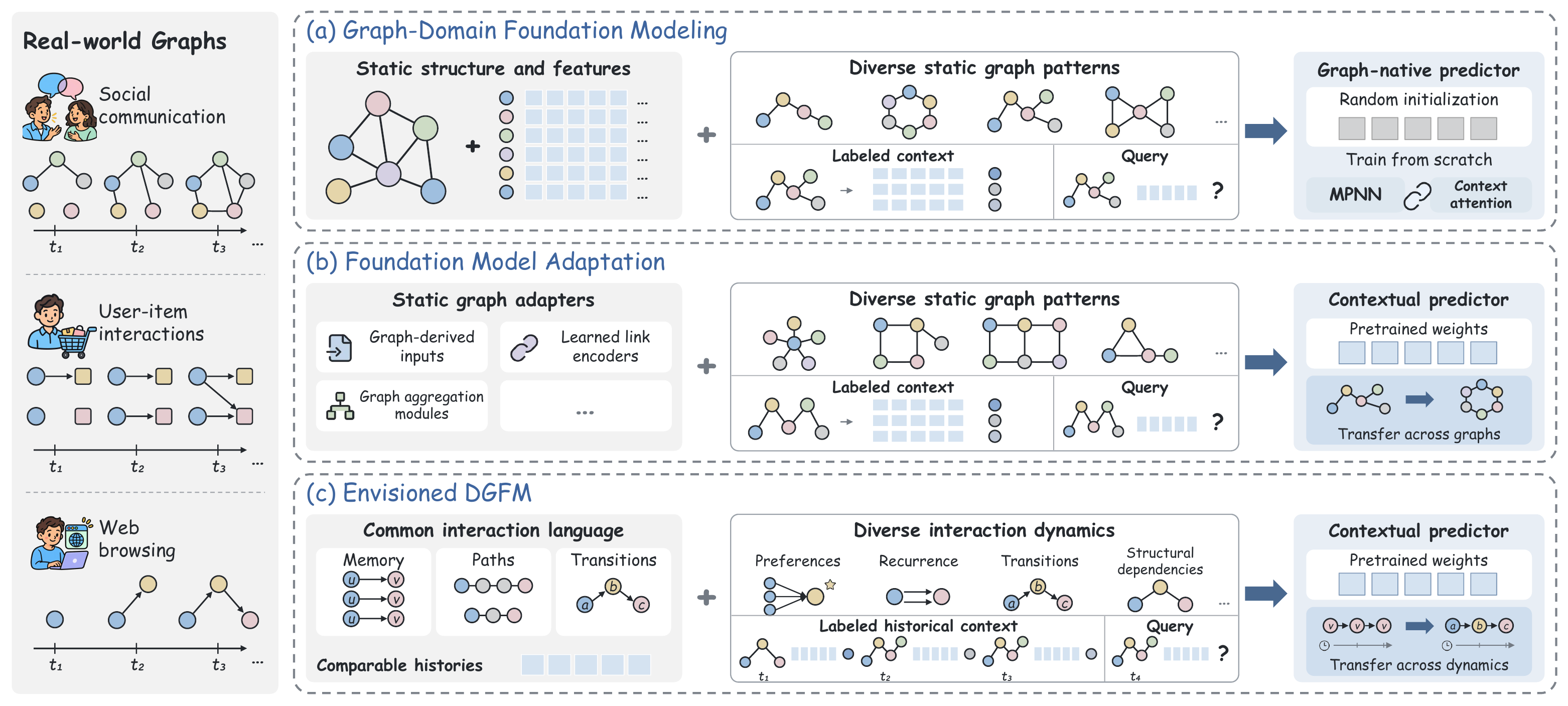}
\vspace{-15pt}
\caption{\textbf{Motivation for \method{}.} (a) Graph-Domain Foundation Modeling couples encoding with prediction. (b) Foundation Model Adaptation in static settings loses temporal distinctions. (c) An envisioned DGFM uses a common interaction language for shared prediction across dynamic graphs.}
\vspace{-20pt}
\label{figure:introduction}
\end{figure}

Research on static graph foundation models has explored how predictive capabilities can be reused across graphs, offering a starting point for extending foundation modeling to dynamic settings. Existing approaches follow two broad routes, illustrated in Fig.~\ref{figure:introduction}. \ding{192}~\textbf{Graph-Domain Foundation Modeling} (Fig.~\ref{figure:introduction}a): Dedicated graph architectures learn transferable prediction rules through graph pretraining. NodePFN combines local message passing with contextual prediction and trains on synthetic graph priors~\cite{choi2026nodepfn}. Its graph encoding is integrated into the predictive architecture, and its static node-level formulation does not model how event histories shape subsequent interactions. \ding{193}~\textbf{Foundation Model Adaptation} (Fig.~\ref{figure:introduction}b): G2T-FM, TFMLinker, and GraphPFN extend existing foundation predictors through graph-derived inputs, learned link encoders, or graph aggregation modules~\cite{eremeev2025g2tfm,liao2026tfmlinker,eremeev2026graphpfn}. These methods demonstrate the value of reusing pretrained predictive knowledge, but their static formulations leave interaction timing and order outside the representation. Early temporal approaches such as TGPM and DyGFM explore transferable patterns in evolving networks through graph-specific pretraining~\cite{ma2026tgpm,yuan2026dygfm}.  A central design question is how to express dynamic histories in a \textbf{common language} and learn predictive relationships that a shared model can apply across different interaction processes.

A common interaction language must describe historical behavior independently of its network. However, node identities are graph-specific, and raw interaction counts and time intervals depend on network activity. This raises \textbf{Challenge~\ding{182}: How can interaction histories be made comparable across dynamic graphs?} For example, repeated contacts within minutes in a highly active network and within days in a less active network may both reflect a persistent relationship, despite their different time intervals. Directly comparing these intervals can obscure the similarity between the two histories. Conversely, equal interaction counts can conceal differences in when contacts occurred, how endpoints were indirectly connected, and in what order destinations were visited. Removing node identities or summarizing histories through counts alone therefore does not establish a common language. Cross-graph comparability requires \emph{accounting for differences in temporal and activity scales while preserving distinctions in interaction timing, historical connectivity, and destination order}.

A common interaction language makes histories comparable across graphs, but a shared model must still learn how to use these histories for prediction. Different networks may rely on different combinations of repeated contacts, historical connectivity, and destination transitions. Learning from a limited range of interaction processes can consequently favor particular historical signals, restricting transfer to networks where other signals matter. This raises \textbf{Challenge~\ding{183}: How can predictive capabilities transfer across networks with different interaction dynamics?} For example, repeated contacts may be informative in networks dominated by persistent relationships, whereas destination transitions may be more useful when interactions follow sequential routines. Strong performance under one process therefore does not ensure reliable prediction under another. Transfer requires \emph{learning to use multiple sources of historical evidence without becoming dependent on a particular interaction process}. This calls for training tasks that cover diverse interaction mechanisms and their combinations, together with historical examples that help relate these signals to observed outcomes in the target network. These challenges motivate the envisioned DGFM in Fig.~\ref{figure:introduction}c: a common interaction language makes histories comparable across networks, while a shared predictor uses historical context to support transfer across different interaction dynamics without target-graph parameter updates.

To realize this vision, we propose \method{}, a Common Language Dynamic Graph Foundation Model for temporal link prediction. For \textbf{Challenge~\ding{182}}, we introduce a \emph{Dynamic Interaction Tokenizer} that encodes interaction memory, historical connectivity, and ordered destination transitions through shared historical descriptors. Temporal and activity calibration support comparability across networks with different identities and scales. These descriptors form feature vectors for conventional machine learning and neural models. Stacking them into tables further enables pretrained tabular foundation models to perform dynamic graph prediction directly, without an additional trainable graph encoder. For \textbf{Challenge~\ding{183}}, \method{} combines \emph{dynamic-prior learning with contextual prediction}. A Dynamic Interaction Prior generates synthetic sequences with varying preferences, recurrence, transitions, and structural dependencies, training an existing foundation predictor across different relationships between histories and future outcomes. Labeled historical context guides prediction on each target graph, with a Contextual Value Adapter adjusting attention values according to the context. Model parameters remain fixed across target graphs. Experiments on TGB and TGB-Seq validate this approach, while comparisons across predictor families demonstrate the common language's broader use beyond the complete \method{} framework.

Our contributions are summarized below:
\begin{itemize}
    \item We introduce \method{}, which reuses a pretrained tabular foundation predictor for temporal link prediction through a common interaction language and synthetic training, without target-graph parameter updates.
    \item We develop a Dynamic Interaction Tokenizer that makes dynamic histories accessible to general-purpose predictors. A Dynamic Interaction Prior and contextual prediction with a Contextual Value Adapter support learning and applying predictive relationships across interaction dynamics.
    \item Experiments on TGB and TGB-Seq show that \method{} achieves the highest mean MRR on 10 of 13 datasets. Further analyses demonstrate that the common language enables direct use of frozen tabular foundation models and competitive prediction with basic machine learning and neural models, while revealing the roles of historical components, training diversity, and context.
\end{itemize}

\section{Related Work}
\label{sec:related_work}

\subsection{Graph Foundation Models}

Graph foundation models transfer predictive knowledge across heterogeneous graphs and tasks. \textbf{Graph-domain foundation modeling} learns shared prediction rules through graph-oriented pretraining. One for All and GIT unify prediction tasks, while UniGraph and GOFA integrate language representations with graph modeling~\cite{liu2024ofa,wang2025git, he2025unigraph,kong2025gofa}. NodePFN combines message passing and context attention through synthetic-prior training~\cite{choi2026nodepfn}. They couple transferable representations with graph-specific architectures. \textbf{Foundation model adaptation} supplies graph
information to pretrained predictors. G2T-FM constructs neighborhood and structural features~\cite{eremeev2025g2tfm}, GraphPFN adds graph aggregation to LimiX and trains on synthetic graphs~\cite{eremeev2026graphpfn}, and TFMLinker uses topology-aware encoders and context prototypes for static link prediction~\cite{liao2026tfmlinker}. Their static formulations omit interaction timing and order.
\method{} extends this route to dynamic graphs via a parameter-free interaction tokenizer and a dynamic interaction prior.

\subsection{Temporal Link Prediction}

\textbf{Memory and neighborhood modeling} extracts historical evidence through node states or temporal aggregation. TGN maintains event-driven memories, while TGAT, GraphMixer, and DyGFormer use temporal attention, mixing, or neighbor co-occurrence~\cite{rossi2020tgn, xu2020tgat,cong2023graphmixer,yu2023dygformer}. 
\textbf{Structural and sequential pattern modeling} captures more explicit dependencies. CAWN and TPNet encode temporal walks, while CRAFT emphasizes frequency patterns and histories conditioned on the destination~\cite{wang2021caw, lu2024tpnet, yi2025craft}.
These models typically require dataset-specific training. For \textbf{transfer across evolving networks}, TGPM pretrains interaction-patch representations, DyGFM separates semantic and temporal representations for multi-domain pretraining, and TIDES combines a frozen time-series foundation encoder with a GNN~\cite{ma2026tgpm,yuan2026dygfm,decastelli2026tides}. Our common interaction language instead supports different predictor families, with \method{} using historical context without target-graph parameter updates.

\subsection{In-Context Learning and Synthetic Priors}

Prior-data fitted networks learn to predict query labels from labeled context using tasks sampled from synthetic priors~\cite{muller2022transformers}. TabPFN and TabICL develop this approach for tabular prediction~\cite{hollmann2025tabpfn,qu2026tabiclv2}, while Mitra explores mixtures of synthetic priors~\cite{zhang2025mitra}. TabDPT instead pretrains on real tables~\cite{ma2025tabdpt}, and LimiX studies a shared conditional prediction interface across structured-data tasks~\cite{zhang2025limix}. Domain transfer also requires suitable input representations~\cite{fang2025benefits,fang2025homophily}. TabPFN-TS formulates forecasting as tabular regression using temporal features, while RDBLearn constructs parameter-free relational representations~\cite{hoo2026tabpfnts,xu2026rdblearn}. For dynamic graphs, \method{} couples a shared representation of memory, connectivity, and event order with synthetic priors that vary their predictive roles.

\section{Problem Formulation}
\label{sec:problem}

\label{sec:problem}

We study how a graph foundation model can perform temporal link prediction across different dynamic graphs. We first define the prediction task and then specify the expected behavior of a shared model on previously unseen networks.

\noindent\textbf{Temporal link prediction.}
A dynamic graph represents interactions between entities over time. We write it as $G=(\mathcal V^G,\mathcal E^G)$, where $\mathcal V^G$ is the node set and $\mathcal E^G=\{(u_i,v_i,t_i)\}_{i=1}^{N_G}$ is the event sequence. Each event records an interaction from source $u_i$ to destination $v_i$ at time $t_i$. Interactions may repeat, and different events may share a timestamp. For a prediction at time $t$, the available history contains only earlier events:
\begin{equation}
\mathcal H^G_{<t}=\{(u_i,v_i,t_i)\in\mathcal E^G \mid t_i<t\},
\label{eq:interaction_history}
\end{equation}
subject to the history-visibility rules of the evaluation protocol.

A query $q=(u,t,\mathcal C_q)$ specifies a source node, a query time, and a candidate destination list $\mathcal C_q=(v_{q,1},\ldots,v_{q,K_q})$. The task is to rank these candidates using the available history, so that the observed destination receives a higher score than alternative destinations. Candidates may be previous interaction partners or nodes that have never interacted directly with the source. Given the source $u$ and query time $t$, the task focuses on predicting the interaction destination from $C_q$.

\noindent\textbf{Foundation modeling across dynamic graphs.}
We seek a graph foundation model that reuses predictive knowledge across networks with different entities, structures, and interaction dynamics. Given the historical graph $G_{<t}=(\mathcal V^G,\mathcal H^G_{<t})$ and a query $q$, the model produces candidate scores:
\begin{equation}
\mathbf s_q^G=\operatorname{GFM}_{\theta}(G_{<t},q)\in\mathbb R^{K_q},
\label{eq:dgfm_prediction}
\end{equation}
where $s_{q,j}^G$ is the score assigned to destination $v_{q,j}$. Sorting these scores gives the predicted destination ranking.
The same parameters $\theta$ are used across target graphs, including previously unseen networks, without target-graph parameter updates. Prediction may use the target graph's historical interactions available under the evaluation protocol, and no correspondence between node identities across graphs is required.

\begin{figure*}[!t]
\centering
\includegraphics[width=\textwidth]{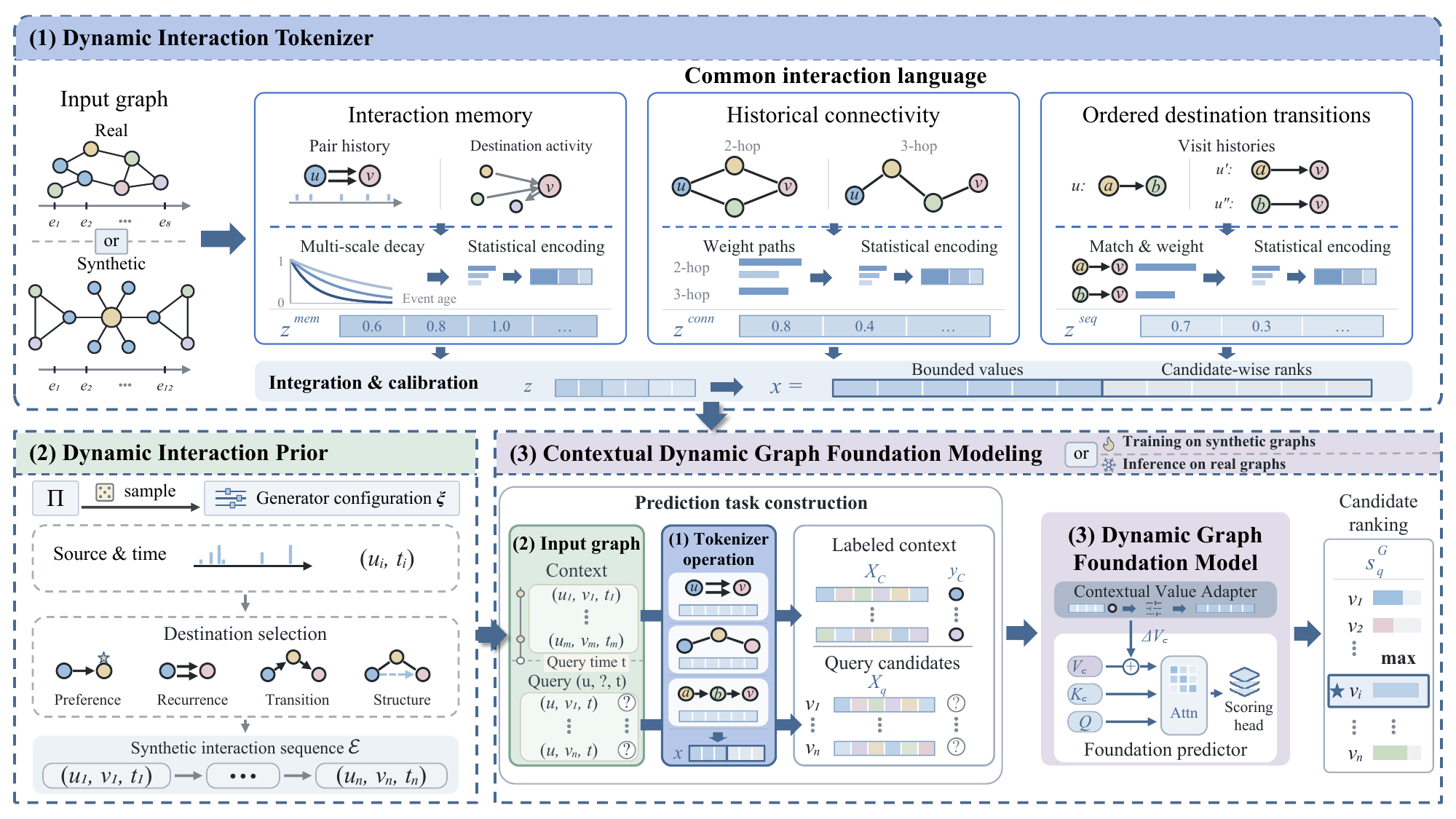}
\vspace{-15pt}
\caption{Overview of \method{}.
(1) The \textbf{Dynamic Interaction Tokenizer} converts real and synthetic histories into a common tabular representation.
(2) The \textbf{Dynamic Interaction Prior} generates diverse interaction dynamics for training.
(3) \textbf{Contextual Dynamic Graph Foundation Modeling} reuses a pretrained tabular foundation predictor, with a Contextual Value Adapter modulating attention values from labeled history.
The model learns on synthetic graphs and ranks candidates on real graphs through historical context, without target-graph parameter updates or an additional trainable graph encoder.}
\label{fig:framework}
\vspace{-15pt}
\end{figure*}


\section{Method}
\label{sec:method}

In this section, we present \method{} (Figure~\ref{fig:framework}). We first construct a common interaction language, then generate training sequences with different history--outcome relationships, and finally train a contextual foundation predictor that uses historical context to predict on target graphs without parameter updates.

\subsection{Dynamic Interaction Tokenizer}
\label{sec:dynamic_tokenizer}

Reusing a predictor across dynamic graphs requires historical evidence with consistent meanings despite different node identities and activity scales. Our tokenizer describes each candidate $(u,v,t)$ through three complementary components: interaction memory captures direct contacts, historical connectivity supplies indirect evidence, and ordered destination transitions capture sequential behavior. All components use only $\mathcal H^G_{<t}$, without encoding node identities. Their outputs are integrated and calibrated into shared numerical inputs for synthetic learning and real-graph prediction.

\subsubsection{Interaction Memory(IM)}

Equal interaction counts can conceal different temporal distributions, while frequent contacts may reflect either pair-specific behavior or high destination activity. We therefore encode pair and destination histories at multiple temporal scales. At query time $t$, let $E_{uv}$ contain timestamps of all previous $u\to v$ interactions, and $I_v$ those of retained recent events involving destination $v$ in either direction. For either timestamp multiset $D$, we define a temporal memory summary using four positive, graph-calibrated scales $h\in\mathcal B$:
\begin{equation}
\mathbf m_D(t)
=
\left[
\log\left(1+\sum_{\tau\in D}e^{-(t-\tau)/h}\right)
\right]_{h\in\mathcal B}.
\label{eq:decayed_memory}
\end{equation}
Here, $\tau$ is an event timestamp. Smaller scales emphasize recent interactions, while larger scales retain older evidence.

Applying this summary to $E_{uv}$ and $I_v$ captures pair-specific and destination activity. We combine them with supplementary summaries $\mathbf a(u,v,t)$ to retain frequency, recency, history availability, and relative pair activity:
\begin{equation}
\mathbf z^{\mathrm{mem}}(u,v,t)
=
\mathbf m_{E_{uv}}(t)
\oplus
\mathbf m_{I_v}(t)
\oplus
\mathbf a(u,v,t),
\label{eq:interaction_memory_representation}
\end{equation}
where $\oplus$ denotes concatenation. 
The vector $\mathbf z^{\mathrm{mem}} \in\mathbb R^{18}$ 
contains eight memory features, three log-counts and three recencies for pair, source, and destination histories, a reverse interaction log-count, observed-pair and unseen-destination indicators, and one normalized memory feature. The latter divides the pair's decayed count by one plus the destination's decayed count at the same selected scale, before log compression. 

\subsubsection{Historical Connectivity(HC)}

Without previous contact, pair memory cannot distinguish candidates by their direct histories. We therefore summarize indirect connections through historical paths, using temporal decay to emphasize recent evidence and activity weighting to discount highly active intermediaries. Let $\mathcal P_{\ell}(u,v,t)$ contain retained length-$\ell$ paths between $u$ and $v$, treating past events as undirected connections without requiring chronological traversal. For path $\rho$, let $\bar a(\rho)$ be its mean event age at $t$, and $w(\rho)$ a unit weight or an inverse-activity weight for intermediaries or the destination. The connectivity vector is:

\begin{equation}
\mathbf z^{\mathrm{conn}}(u,v,t)
=
\left[
\log\left(
1+\sum_{\rho\in\mathcal P_{\ell}(u,v,t)}
w(\rho)e^{-\bar a(\rho)/h}
\right)
\right]_{(\ell,w,h)\in\Omega_P}.
\label{eq:historical_connectivity_vector}
\end{equation}
Here, $\Omega_P$ specifies path lengths, weights, and decay scales $h$. With $\ell\in\{2,3\}$, $\mathbf z^{\mathrm{conn}}(u,v,t)\in\mathbb R^7$ records six connectivity statistics for paths of lengths two and three and one three-step count normalized by destination activity through $w(\rho)$.

\subsubsection{Ordered Destination Transitions(ODT)}

Static path connectivity ignores the sequential order in which destinations are visited. We complement it with collaborative destination transitions observed across other sources, capturing which destinations typically follow $u$'s recent visits. We consider the destinations of $u$'s latest $L$ outgoing events and look for other sources that visited these destinations and subsequently visited candidate $v$. Each such pair of visits forms a transition occurrence, with both visits strictly before $t$. Let $\mathcal O_L(u,v,t)$ collect retained occurrences, counted separately for each of these outgoing events.

Transitions are more informative when they relate to $u$'s more recent visits or occur earlier among another source's subsequent visits. We encode these preferences in a weight $\omega(e)$ for each occurrence $e$, optionally downweighting highly active sources. To favor closely spaced visits, we decay each contribution by the interval $\Delta(e)>0$ between the two visits, using a time scale $h>0$. For the current query $(u,v,t)$, we aggregate this evidence into a destination-transition score:
\begin{equation}
T_L(\omega,h)
=
\sum_{e\in\mathcal O_L(u,v,t)}
\omega(e)e^{-\Delta(e)/h}.
\label{eq:destination_transition_score}
\end{equation}

Using $L\in\{1,3\}$ captures evidence from the latest visit and a broader recent history. Let $\Omega_T$ collect the settings $(L,\omega,h)$, including destination-activity normalization through $\omega$. We log-compress the scores and append an indicator of immediate repetition, where $v_u^{\mathrm{last}}$ denotes $u$'s latest destination:
\begin{equation}
\mathbf z^{\mathrm{seq}}(u,v,t)
=
\left[
\log\left(1+T_L(\omega,h)\right)
\right]_{(L,\omega,h)\in\Omega_T}
\oplus
[\mathbf1\{v=v_u^{\mathrm{last}}\}].
\label{eq:ordered_transition_vector}
\end{equation}
The vector $\mathbf z^{\mathrm{seq}}(u,v,t)\in\mathbb R^6$ combines four transition statistics capturing reference recency, time gaps, and source activity with a destination-activity-normalized count and the latest-destination indicator. The indicator is zero without outgoing history.

\paragraph{Input integration and calibration.}

We concatenate the three complementary representations into a shared history descriptor:
\begin{equation}
\mathbf z(u,v,t)
=
\mathbf z^{\mathrm{mem}}(u,v,t)
\oplus
\mathbf z^{\mathrm{conn}}(u,v,t)
\oplus
\mathbf z^{\mathrm{seq}}(u,v,t)
\in\mathbb R^{31}.
\label{eq:integrated_history}
\end{equation}

To limit feature magnitudes while retaining their relative ordering within each candidate list, we combine bounded values with candidate-wise ranks. For query $q$ at source $u$ and time $t$, write $\mathbf z_{q,j}=\mathbf z(u,v_{q,j},t)$ for its $j$th candidate among $K_q$ candidates. Let $\mathbf r_{q,j}$ contain its coordinatewise ascending ranks within this list, using average ranks for ties and rescaling into $(-1,1)$. With compression scale $c_x>0$, the final candidate input is:
\begin{equation}
\mathbf x_{q,j}
=
\tanh(\mathbf z_{q,j}/c_x)
\oplus
\mathbf r_{q,j}
\in\mathbb R^{62}.
\label{eq:tokenizer_output}
\end{equation}
Stacking these vectors as rows gives the query input $X_q\in\mathbb R^{K_q\times p}$, where $p=62$. The same label-free transformations construct historical context inputs, with ranks computed within each original candidate list before context-row selection. Time-scale calibration and numerical settings are specified in Appendix~\ref{app:tokenizer}.

\subsection{Dynamic Interaction Prior}
\label{sec:dynamic_prior}

The tokenizer establishes shared historical descriptors, but their predictive effects vary across interaction processes. Repetition may reinforce future contact under persistent preferences or suppress it under repetition inhibition. Our Dynamic Interaction Prior varies these mechanisms so that the foundation predictor learns to interpret historical evidence using labeled context. We draw generator configurations $\xi$ from a prior $\Pi$, varying source activity and destination-selection mechanisms. These configurations remain hidden from the predictor.

\subsubsection{Source Activity and Event Timing}

Different activity patterns yield different amounts and temporal distributions of historical evidence. We first generate event times and sources using exponential Hawkes processes~\cite{hawkes1971spectra}. For each source $u$, the event rate $\lambda_u(t)$ combines baseline activity with exponentially decaying excitation from its past events, producing heterogeneous activity and bursts.

For event $i$, the total source rate governs the next event time $t_i$. At $t_i$, we sample source $u_i$ proportionally to rates based on events strictly before $t_i$, then sample destination $v_i$ from the conditional distribution $\pi_{\xi}(\cdot\mid u_i,t_i)$ defined next.

\subsubsection{History-Dependent Destination Selection}

We combine preference, recurrence, transition, and structural mechanisms into a destination-sampling distribution, allowing different historical signals and preferences to govern the next interaction. 
\textbf{Preference} scores $g_{\mathrm{pref}}$ capture source--destination affinity and destination popularity, allowing interactions between pairs with no previous contact. 
\textbf{Recurrence} scores $g_{\mathrm{rec}}$ instead sum exponentially decayed past $u\to v$ interactions, favoring repeated direct contacts.

The other two mechanisms condition on $u$'s latest destination $v_u^{\mathrm{last}}$.
\textbf{Transition} scores $g_{\mathrm{seq}}$ use a sampled distribution $Q(v\mid a)$ specifying preferences for visiting $v$ after destination $a$.
\textbf{Structural} scores $g_{\mathrm{str}}$ draw on a sampled set $\mathcal S_u(t)$ of other sources that previously visited $v_u^{\mathrm{last}}$.
Let $C_{wv}(t)$ count past $w\to v$ events and $n_w^{\mathrm{out}}(t)$ count all past outgoing events of $w$. These scores are:
\begin{equation}
\begin{aligned}
g_{\mathrm{seq}}(u,v,t)
&=Q(v\mid v_u^{\mathrm{last}}),\\
g_{\mathrm{str}}(u,v,t)
&=\sum_{w\in\mathcal S_u(t)}
\frac{C_{wv}(t)}{1+n_w^{\mathrm{out}}(t)}.
\end{aligned}
\label{eq:generative_mechanism_scores}
\end{equation}
Thus, closure favors destinations visited by these other sources while discounting highly active sources.

For $m\in\{\mathrm{pref},\mathrm{rec},\mathrm{seq},\mathrm{str}\}$, let $\bar g_m$ denote $g_m$ normalized to sum to one over destinations. We mix these scores with nonnegative weights $\alpha_m$ summing to one. To allow repeated contacts to become less likely despite recurrence, we penalize previous pair interactions with strength $\gamma\geq0$:
\begin{equation}
\pi_{\xi}(v\mid u,t)
\propto
e^{-\gamma C_{uv}(t)}
\sum_m\alpha_m\bar g_m(u,v,t).
\label{eq:destination_prior}
\end{equation}

We normalize over valid destinations and draw $v_i\sim\pi_\xi(\cdot\mid u_i,t_i)$, completing the event $(u_i,v_i,t_i)$. Repeating this procedure and updating the history after each event produces a sequence $\mathcal E$. Varying the generator configuration yields sequences with different history--outcome relationships, from which Section~\ref{sec:foundation_model} constructs training tasks. Definitions and variants of $g_{\mathrm{pref}}$ and $g_{\mathrm{rec}}$, sampling and fallback rules for $g_{\mathrm{seq}}$ and $g_{\mathrm{str}}$, and numerical details of $\pi_\xi$ are provided in Appendix~\ref{app:dynamic-prior}.

\subsection{Contextual Dynamic Graph Foundation Modeling}
\label{sec:foundation_model}

A common representation alone does not determine which historical patterns predict future interactions. We use labeled historical context to interpret the shared inputs from Section~\ref{sec:dynamic_tokenizer}. Training tasks constructed from the sequences generated in Section~\ref{sec:dynamic_prior} expose the predictor to different relationships between historical evidence and future interactions.

\subsubsection{History-Conditioned Prediction}

We build \method{} on a pretrained TabICLv2 backbone~\cite{qu2026tabiclv2}, whose in-context predictor uses labeled historical examples to score query candidates. For earlier events, we pair observed destinations with sampled alternatives, labeled one and zero, respectively. Applying Section~\ref{sec:dynamic_tokenizer} yields a context table $X_C\in\mathbb R^{n_C\times62}$ and labels $\mathbf y_C\in\{0,1\}^{n_C}$, where $n_C$ is the number of retained rows. For query $q$, the same tokenizer produces $X_q\in\mathbb R^{K_q\times62}$ for its $K_q$ candidates. Each row uses only history preceding its event time.

Historical features can have different predictive meanings across tasks. We introduce a Contextual Value Adapter that uses labeled context to adjust the content transmitted by context rows. Its inputs are the existing feature values in $X_C$, their labels $\mathbf y_C$, and the backbone's context-row representations. Statistics of positive and negative context rows guide memory-scale combinations and gate a low-rank value correction. Let $\bar H_C$ contain the context representations after label embedding and normalization, and let $Q$, $K_C$, and $V_C$ denote the first attention layer's projected queries, context keys, and context values. The adapter modifies attention as follows:
\begin{equation}
\begin{aligned}
\Delta V_C
&=
\operatorname{ValueAdapter}
(\bar H_C;X_C,\mathbf y_C),\\
O
&=
\operatorname{Attn}
(Q,K_C,V_C+\Delta V_C).
\end{aligned}
\label{eq:contextual_value_modulation}
\end{equation}
Here, $\Delta V_C$ is the correction in projected value space. The backbone's multi-head attention operator $\operatorname{Attn}$ uses queries from context and query rows, with keys and values only from context rows.

The attention output $O$ continues through the backbone, whose scoring head maps the resulting query representations to candidate scores. Let $F_\theta$ denote the complete predictor, with $\theta$ collecting the trainable backbone and adapter parameters. We obtain the ranking vector defined in Eq.~\eqref{eq:dgfm_prediction}:
\begin{equation}
\mathbf s_q^G
=
F_\theta(X_C,\mathbf y_C,X_q)
\in\mathbb R^{K_q}.
\label{eq:contextual_candidate_scores}
\end{equation}
Its $j$th entry scores candidate $v_{q,j}$, with larger values indicating stronger predictions. The training objective below uses these scores. Context selection and adapter details are in Appendix~\ref{app:contextual-modeling}.

\subsubsection{Dynamic-Prior Learning and Cross-Graph Inference}

From a sequence $\mathcal E$ generated in Section~\ref{sec:dynamic_prior}, we construct a prediction task $\mathcal T$ using earlier events as labeled context and later events as queries, following the input construction above. Let $\mathcal D_\Pi$ denote the task distribution induced by sequence generation and this construction. Variation across these sequences encourages the predictor to learn how context determines the predictive roles of historical features.

Starting from the pretrained backbone, we freeze its input encoders and jointly train the in-context predictor and Contextual Value Adapter. For a task $\mathcal T\sim\mathcal D_\Pi$, let $G$ be its underlying graph, $\mathcal I_Q$ its query set, and $j_q^\star$ the observed destination's candidate index. Using the scores from Eq.~\eqref{eq:contextual_candidate_scores}, we minimize a listwise ranking loss:
\begin{equation}
\mathcal L(\theta)
=
-\mathbb E_{\mathcal T\sim\mathcal D_\Pi}
\left[
\frac{1}{|\mathcal I_Q|}
\sum_{q\in\mathcal I_Q}
\log
\frac{\exp(s_{q,j_q^\star}^{G})}
{\sum_{j=1}^{K_q}\exp(s_{q,j}^{G})}
\right].
\label{eq:dynamic_training}
\end{equation}
Here, $s_{q,j}^{G}$ is the $j$th entry of $\mathbf s_q^G$. The objective favors the observed destination over alternative candidates, using interaction outcomes rather than generator parameters as supervision.

On a real graph $G$, we construct labeled context from its training interval and represent each later query using history available before its timestamp. The same predictor produces $\mathbf s_q^G$ for candidate ranking, ordering candidates by score and selecting the top-ranked destination as the predicted link target. All learned parameters remain fixed, while the graph's labeled context conditions both value modulation and in-context prediction. Task sampling, optimization, and inference details are provided in Appendix~\ref{app:contextual-modeling}.

\section{Experiments}
\label{sec:experiments}

Our experiments address three research questions:
\textbf{RQ1:} How effectively does \method{} predict temporal links across graphs without target-graph parameter updates?
\textbf{RQ2:} How effective is the Dynamic Interaction Tokenizer as a common representation across graphs and predictors?
\textbf{RQ3:} How do interaction diversity during training and historical context support cross-graph transfer?

\subsection{Experimental Setup}
\label{sec:experimental-setup}

We compare with 16 temporal link prediction baselines.
\textbf{Memory and neighborhood modeling} covers EdgeBank~\cite{poursafaei2022evaluation}, DyRep~\cite{trivedi2019dyrep}, JODIE~\cite{kumar2019jodie}, TGN~\cite{rossi2020tgn}, TGAT~\cite{xu2020tgat}, TCL~\cite{wang2021tcl}, GraphMixer~\cite{cong2023graphmixer}, DyGFormer~\cite{yu2023dygformer}, CTAN~\cite{gravina2024ctan}, and TSNN~\cite{asif2026tsnn}, which use edge memorization, evolving node states, or temporal neighborhood aggregation for prediction.
\textbf{Structural and sequential pattern modeling} covers CAWN~\cite{wang2021caw}, TNCN~\cite{zhang2025tncn}, TPNet~\cite{lu2024tpnet}, HyperEvent~\cite{gao2025hyperevent}, CRAFT~\cite{yi2025craft}, and TAMI~\cite{yu2025tami}, which explicitly represent temporal connectivity and dependencies within interaction histories.

To assess predictor compatibility, we additionally compare ten predictors using the same tokenizer.
\textbf{Machine learning} models include logistic regression, random forest, CatBoost~\cite{prokhorenkova2018catboost}, and LightGBM~\cite{ke2017lightgbm}.
\textbf{Deep learning} models include MLP, ResNet, and FT-Transformer~\cite{gorishniy2021revisiting}.
\textbf{Pretrained foundation models} include TabICLv2~\cite{qu2026tabiclv2}, TabPFNv2~\cite{hollmann2025tabpfn}, and LimiX-2M~\cite{zhang2025limix}.
Baseline sources and settings are provided in Appendix~\ref{app:experimental-protocol}.

\noindent\textbf{Dataset.}
We evaluate on 5 datasets from TGB~\cite{huang2023tgb} and 8 from TGB-Seq~\cite{yi2025tgbseq}, covering online interactions, recommendation, financial transactions, transportation, citations, and hyperlinks.

\noindent\textbf{Evaluation.} We evaluate on complete official test sets with candidates from the benchmark and report $100\times\mathrm{MRR}$.
Trained exclusively on synthetic interaction streams, \method{} uses the same parameters across all target graphs, conditioning on strictly earlier interactions and fixed labeled context from each graph's training split without using native node or edge attributes.
Learned temporal baselines are trained and validated separately on each graph.
We report the mean and sample standard deviation of \method{} over three training seeds.
\textbf{Dataset}, \textbf{evaluation}, and \textbf{implementation details} are provided in Appendices~\ref{app:experimental-protocol} and~\ref{app:contextual-modeling}.

\definecolor{SeqGroup}{HTML}{F3EFE7}
\definecolor{SeqOurs}{HTML}{E2EFFB}
\definecolor{SeqAccent}{HTML}{254E70}
\definecolor{SeqBest}{HTML}{B32632}

\begin{table}[!t]
\centering\small
\setlength{\tabcolsep}{3pt}
\renewcommand{\arraystretch}{1.08}

\newcommand{\seqval}[2]{%
    \ensuremath{#1_{\pm #2}}%
}
\newcommand{\seqbest}[2]{%
    \seqval{\textcolor{SeqBest}{\mathbf{#1}}}{#2}%
}
\newcommand{\seqsecond}[2]{%
    \seqval{\underline{#1}}{#2}%
}

\caption{Performance on TGB ($100\times\mathrm{MRR}$; higher is better). Best and second-best results per column are \textcolor{SeqBest}{\textbf{bold red}} and \underline{underlined}, respectively.}
\label{tab:main-tgb}
\vspace{-5pt}
\begin{tabular}{@{}lccccc@{}}
\toprule
\textbf{Method} & \textbf{Wiki} & \textbf{Review} & \textbf{Coin} & \textbf{Flight} & \textbf{Comment} \\
\midrule
CAWN & $71.10_{\pm 0.60}$ & $19.30_{\pm 0.10}$ & -- & -- & -- \\
CTAN & $66.80_{\pm 0.70}$ & $40.50_{\pm 0.40}$ & $74.80_{\pm 0.40}$ & -- & $67.10_{\pm 6.70}$ \\
DyGFormer & $79.80_{\pm 0.40}$ & $22.40_{\pm 1.50}$ & $75.20_{\pm 0.40}$ & -- & $67.00_{\pm 0.10}$ \\
DyRep & $5.00_{\pm 1.70}$ & $22.00_{\pm 3.00}$ & $45.20_{\pm 4.60}$ & $55.60_{\pm 1.40}$ & $28.90_{\pm 3.30}$ \\
EdgeBank & $57.10$ & $2.50$ & $58.00$ & $38.70$ & $14.90$ \\
GraphMixer & $11.80_{\pm 0.20}$ & \seqsecond{52.10}{1.50} & -- & -- & -- \\
HyperEvent & $81.00_{\pm 0.20}$ & -- & $77.30_{\pm 0.20}$ & $87.70_{\pm 0.30}$ & $76.00_{\pm 0.20}$ \\
TGAT & $14.10_{\pm 0.70}$ & $35.50_{\pm 1.20}$ & -- & -- & -- \\
TGN & $39.60_{\pm 6.00}$ & $34.90_{\pm 2.00}$ & $58.60_{\pm 3.70}$ & $70.50_{\pm 2.00}$ & $37.90_{\pm 2.10}$ \\
TNCN & $71.80_{\pm 0.10}$ & $37.70_{\pm 1.00}$ & $76.20_{\pm 0.40}$ & $82.00_{\pm 0.40}$ & $69.70_{\pm 0.60}$ \\
TPNet & \seqbest{82.70}{0.10} & -- & \seqsecond{83.20}{0.10} & \seqsecond{88.40}{0.10} & \seqsecond{82.50}{0.60} \\
\midrule

\rowcolor{DGFMHighlight}
\textbf{\method{} } & \seqsecond{82.24}{0.51} & \seqbest{52.89}{0.20} & \seqbest{84.95}{0.06} & \seqbest{90.68}{0.05} & \seqbest{93.28}{0.16} \\

\bottomrule
\end{tabular}
\vspace{-15pt}
\end{table}

\begin{table*}[!t]
\centering
\begingroup
\small
\setlength{\tabcolsep}{3pt}
\renewcommand{\arraystretch}{1.16}
\setlength{\aboverulesep}{0pt}
\setlength{\belowrulesep}{0pt}

\newcommand{\seqval}[2]{%
    \smash{\begingroup\normalsize
    \ensuremath{#1_{\pm #2}}%
    \endgroup}%
}
\newcommand{\seqbest}[2]{%
    \seqval{\textcolor{SeqBest}{\mathbf{#1}}}{#2}%
}
\newcommand{\seqsecond}[2]{%
    \seqval{\underline{#1}}{#2}%
}

\caption{Performance on TGB-Seq ($100\times\mathrm{MRR}$; higher is better). Baselines train per target graph; CL-DGFM uses a shared model with labeled context and fixed parameters. Each column's best result is \textcolor{SeqBest}{\textbf{bold red}} and its second best is \underline{underlined}.}
\label{tab:main-seq}
\vspace{-5pt}
\begin{tabularx}{\textwidth}{
    @{\hspace{5pt}}
    l
    *{8}{>{\centering\arraybackslash}X}
    @{\hspace{5pt}}
}
\toprule[1pt]
\rule{0pt}{3.0ex}
\textbf{Method}
& \textbf{ML-20M}
& \textbf{Taobao}
& \textbf{Yelp}
& \textbf{GoogleLocal}
& \textbf{Flickr}
& \textbf{YouTube}
& \textbf{Patent}
& \textbf{WikiLink}
\\[4pt]
\midrule[0.6pt]

\rowcolor{SeqGroup}
\multicolumn{9}{@{\hspace{5pt}}l@{\hspace{5pt}}}{
    \rule{0pt}{2.6ex}
    \textbf{Memory and neighborhood modeling}
    \hspace{0.7em}
    {\footnotesize\textit{Training per target graph}}
}\\[2pt]

DyRep
& \seqval{19.00}{1.69}
& \seqval{40.03}{2.40}
& \seqval{57.69}{1.05}
& \seqval{37.73}{1.34}
& \seqval{38.04}{4.19}
& \seqval{35.12}{4.13}
& \seqval{21.01}{1.14}
& \seqval{42.63}{1.33} \\

JODIE
& \seqval{21.16}{0.73}
& \seqval{48.36}{2.18}
& \seqval{69.88}{0.31}
& \seqval{41.86}{1.49}
& \seqval{46.21}{0.83}
& \seqval{41.67}{2.86}
& \seqbest{24.60}{0.38}
& \seqval{57.94}{1.33} \\

TGN
& \seqval{23.99}{0.20}
& \seqval{60.28}{0.54}
& \seqval{69.79}{0.24}
& \seqval{54.13}{1.97}
& \seqval{46.03}{6.78}
& \seqval{55.16}{5.89}
& \seqsecond{22.83}{2.25}
& \seqval{62.94}{2.16} \\

TGAT
& \seqval{10.47}{0.20}
& --
& --
& \seqval{19.78}{0.24}
& \seqval{23.53}{3.35}
& \seqval{43.56}{2.53}
& \seqval{8.49}{0.18}
& -- \\

TCL
& \seqval{12.04}{0.02}
& \seqval{31.55}{0.03}
& \seqval{24.39}{0.67}
& \seqval{18.30}{0.02}
& \seqval{40.00}{1.76}
& \seqval{50.17}{1.98}
& \seqval{10.60}{1.75}
& \seqval{43.02}{2.16} \\

GraphMixer
& \seqval{21.97}{0.17}
& \seqval{31.54}{0.02}
& \seqval{33.96}{0.19}
& \seqval{21.31}{0.14}
& \seqval{45.01}{0.08}
& \seqval{58.87}{0.12}
& \seqval{18.97}{2.54}
& \seqval{48.57}{0.02} \\

DyGFormer
& --
& --
& \seqval{21.68}{0.20}
& \seqval{18.39}{0.02}
& \seqval{49.58}{2.87}
& \seqval{46.08}{3.44}
& \seqval{14.20}{2.93}
& -- \\

TSNN
& \seqsecond{35.47}{0.05}
& \seqval{68.25}{0.17}
& \seqval{69.13}{0.29}
& \seqbest{61.43}{0.59}
& \seqval{60.84}{0.13}
& \seqsecond{59.78}{0.32}
& \seqval{17.21}{0.16}
& \seqval{69.37}{0.22} \\[3pt]

\rowcolor{SeqGroup}
\multicolumn{9}{@{\hspace{5pt}}l@{\hspace{5pt}}}{
    \rule{0pt}{2.6ex}
    \textbf{Structural and sequential pattern modeling}
    \hspace{0.7em}
    {\footnotesize\textit{Training per target graph}}
}\\[2pt]

CAWN
& \seqval{12.31}{0.02}
& --
& \seqval{25.71}{0.09}
& \seqval{18.26}{0.02}
& \seqval{48.69}{6.08}
& \seqval{47.55}{1.08}
& \seqval{12.34}{0.47}
& -- \\

CRAFT
& \seqval{34.42}{0.68}
& \seqsecond{69.13}{0.45}
& \seqsecond{71.12}{0.58}
& \seqval{60.81}{0.49}
& \seqsecond{60.87}{0.81}
& \seqval{57.39}{0.19}
& \seqval{18.48}{0.72}
& \seqsecond{73.92}{0.86} \\

TAMI
& \seqval{31.58}{0.03}
& \seqval{67.11}{0.24}
& \seqval{67.80}{0.41}
& \seqsecond{61.35}{0.48}
& \seqval{58.62}{0.22}
& \seqval{58.11}{0.20}
& \seqval{16.53}{0.35}
& \seqval{67.89}{0.42} \\[3pt]

\midrule[0.6pt]

\rowcolor{SeqOurs}
\multicolumn{9}{@{\hspace{5pt}}l@{\hspace{5pt}}}{
    \rule{0pt}{2.8ex}
    \textcolor{SeqAccent}{\textbf{Dynamic graph foundation model}}
    \hspace{0.7em}
    {\footnotesize\textit{One shared model; in-context prediction; no target-graph fine-tuning}}
}\\[2pt]

\rowcolor{SeqOurs}
\rule{0pt}{2.6ex}\textbf{CL-DGFM}
& \seqbest{36.64}{0.63}
& \seqbest{69.67}{0.03}
& \seqbest{72.48}{0.08}
& \seqval{57.36}{0.23}
& \seqbest{76.11}{0.21}
& \seqbest{71.47}{0.01}
& \seqval{19.32}{0.61}
& \seqbest{74.25}{0.17} \\[4pt]

\bottomrule[1pt]
\end{tabularx}
\endgroup
\vspace{-10pt}
\end{table*}

\subsection{Cross-Graph Prediction Performance (RQ1)}

We evaluate whether a common interaction language enables shared prediction across networks with different interaction dynamics. Tables~\ref{tab:main-tgb} and~\ref{tab:main-seq} report the results.

\noindent\textbf{Obs. 1. A common interaction language supports shared prediction across dynamic graphs.}
\method{} achieves the highest mean MRR on 10 of 13 datasets. It exceeds the strongest listed baselines by 10.78, 15.24, and 11.69 percentage points on Comment, Flickr, and YouTube, respectively. These results support the route proposed in the introduction: \textbf{shared historical descriptors and learning across diverse synthetic dynamics enable an existing foundation predictor to transfer across real networks}. Labeled historical context conditions prediction without target-graph parameter updates or an extra trainable graph encoder. \method{} remains slightly below TPNet on Wiki and shows larger gaps on GoogleLocal and Patent. Shared descriptors may omit useful historical details, while the synthetic prior and limited context may not sufficiently cover some target dynamics. This suggests that a common representation supports transfer, but effectiveness depends on learning and identifying relevant history-outcome relationships.

\definecolor{PredGroup}{HTML}{F3EFE7}
\definecolor{PredFrozen}{HTML}{EFF5FA}
\definecolor{PredOurs}{HTML}{E2EFFB}
\definecolor{PredAccent}{HTML}{254E70}
\definecolor{PredBest}{HTML}{B32632}

\begin{table}[t]
\centering
\begingroup
\small
\setlength{\tabcolsep}{2pt}
\renewcommand{\arraystretch}{1.12}
\setlength{\aboverulesep}{0pt}
\setlength{\belowrulesep}{0pt}

\newcommand{\predbest}[1]{%
    \ensuremath{\textcolor{PredBest}{\mathbf{#1}}}%
}
\newcommand{\predsecond}[1]{%
    \ensuremath{\underline{#1}}%
}

\caption{Predictor comparison using the same tokenizer
($100\times\mathrm{MRR}$; higher is better).
Column-wise best results are
\textcolor{PredBest}{\textbf{red and bold}};
second-best results are \underline{underlined}.}
\label{tab:analysis-predictors}
\vspace{-5pt}
\begin{tabularx}{\columnwidth}{
    @{\hspace{3pt}}
    l
    *{4}{>{\centering\arraybackslash}X}
    @{\hspace{3pt}}
}
\toprule[0.9pt]
\rule{0pt}{2.8ex}
\textbf{Predictor}
& \textbf{Wiki}
& \begin{tabular}[c]{@{}c@{}}\textbf{Google}\\\textbf{Local}\end{tabular}
& \textbf{Flickr}
& \textbf{YouTube}
\\[3pt]
\midrule[0.5pt]

\rowcolor{PredGroup}
\multicolumn{5}{@{\hspace{3pt}}l@{\hspace{3pt}}}{
    \rule{0pt}{2.5ex}
    \textbf{Machine learning}
    \hspace{0.5em}
    {\footnotesize\textit{Training per graph}}
}\\[1pt]

Logistic regression
& $79.73$ & $54.31$ & $70.07$ & $69.32$ \\

Random forest
& $79.25$ & $56.10$ & $71.08$ & $69.51$ \\

CatBoost
& $80.65$ & $56.13$ & $72.20$ & $69.95$ \\

LightGBM
& $81.21$ & $54.12$ & $71.94$ & $66.63$ \\[2pt]

\rowcolor{PredGroup}
\multicolumn{5}{@{\hspace{3pt}}l@{\hspace{3pt}}}{
    \rule{0pt}{2.5ex}
    \textbf{Deep learning}
    \hspace{0.5em}
    {\footnotesize\textit{Training per graph}}
}\\[1pt]

MLP
& $80.37$ & $54.07$ & $70.87$ & $69.19$ \\

ResNet
& $81.76$ & $54.56$ & $70.31$ & $67.62$ \\

FT-Transformer
& \predsecond{81.94}
& $55.70$
& \predsecond{75.07}
& $69.74$ \\[2pt]

\rowcolor{PredFrozen}
\multicolumn{5}{@{\hspace{3pt}}l@{\hspace{3pt}}}{
    \rule{0pt}{2.5ex}
    \textcolor{PredAccent}{\textbf{Pretrained foundation models}}
    \hspace{0.5em}
    {\footnotesize\textit{No fine-tuning}}
}\\

\rowcolor{PredFrozen}
\multicolumn{5}{@{\hspace{3pt}}l@{\hspace{3pt}}}{
    {\footnotesize
    Fixed weights across graphs; graph-specific labeled context}
}\\[2pt]

TabICLv2
& $77.19$
& \predsecond{56.63}
& $73.63$
& \predsecond{71.16} \\

TabPFNv2
& $81.12$ & $56.14$ & $74.83$ & $69.00$ \\

LimiX-2M
& $79.82$ & $56.41$ & $72.72$ & $69.91$ \\[2pt]

\cmidrule[0.4pt]{1-5}

\rowcolor{PredOurs}
\rule{0pt}{2.6ex}\textbf{CL-DGFM}
& \predbest{82.24}
& \predbest{57.36}
& \predbest{76.11}
& \predbest{71.47}
\\[3pt]

\bottomrule[0.9pt]
\end{tabularx}
\endgroup
\vspace{-15pt}
\end{table}

\begin{figure}[t]
\centering
\includegraphics[width=0.9\linewidth]{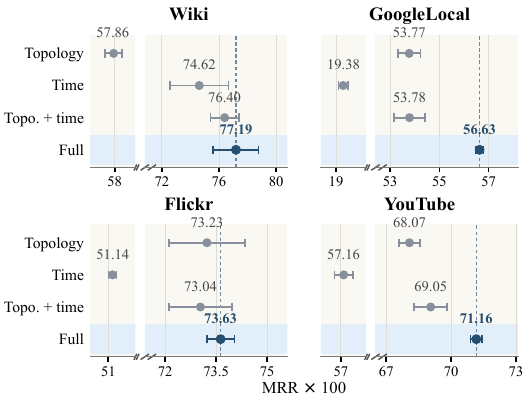}
\vspace{-15pt}
\caption{Representation comparison with a frozen backbone. Error bars show standard deviations over three paired context draws; dashed lines mark the full tokenizer.}
\label{fig:tokenizer-statistics}
\vspace{-15pt}
\end{figure}

\subsection{Effectiveness of the Common Interaction Language (RQ2)}

We represent each candidate interaction as a feature vector and stack these vectors into a table. Using this shared input format, we evaluate conventional machine learning models, deep learning models, and pretrained tabular foundation models. We also compare alternative representations using the same frozen backbone and assess tokenizer components by removing each component while keeping model weights fixed.

\noindent\textbf{Obs. 2. The common language brings tabular foundation models to dynamic graph prediction.}
Table~\ref{tab:analysis-predictors} shows that directly
using tokenizer outputs as feature vectors allows ordinary machine learning and neural models to predict temporal links with competitive performance. More importantly, arranging these outputs as tabular data enables \textbf{direct use of pretrained tabular foundation models on dynamic graphs}. TabICLv2, TabPFNv2, and LimiX-2M perform prediction with fixed parameters and labeled historical context, without an additional graph encoder. Thus, the common language makes dynamic graph prediction accessible to existing predictors and allows \textbf{capabilities learned from tabular tasks to be reused on interaction histories}.

\noindent\textbf{Obs. 3. Explicit interaction patterns outperform combined topology and temporal statistics.}
Figure~\ref{fig:tokenizer-statistics} shows that the complete tokenizer outperforms topology statistics, temporal statistics, and their concatenation on all four graphs. Merely combining topology and time does not consistently improve prediction. The tokenizer instead describes how contacts repeat, how endpoints connect indirectly, and which destinations follow earlier visits. These explicit interaction patterns provide more useful evidence to the same frozen predictor, showing that \textbf{the benefit comes from how histories are represented, beyond a change in input format}.

\begin{figure}[t]
\centering
\includegraphics[width=\linewidth]{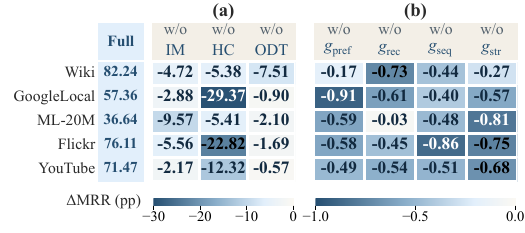}
\vspace{-15pt}
\caption{MRR changes from ablating (a) tokenizer components (RQ2) and (b) prior mechanisms (RQ3).}
\label{fig:tokenizer-ablation}
\vspace{-15pt}
\end{figure}

\noindent\textbf{Obs. 4. Different graphs rely on different combinations of memory, connectivity, and transitions.}
Figure~\ref{fig:tokenizer-ablation}(a) shows that removing connectivity causes the largest drops on GoogleLocal, Flickr, and YouTube. \textbf{Indirect paths can distinguish candidates with little or no direct contact}, which pair memory alone cannot resolve. Memory matters most on ML-20M, consistent with \textbf{repeated contacts and their recency providing useful evidence}. Transitions matter most on Wiki, suggesting that \textbf{destination order adds information beyond contact frequency and connectivity}. These patterns explain the components' complementary roles: a common language must preserve multiple historical signals to support graphs with different interaction dynamics.

\subsection{Factors Supporting Cross-Graph Transfer (RQ3)}

We ablate prior mechanisms with matched retraining budgets, vary the number of distinct synthetic tasks at a fixed budget, and compare predictors across context sizes. These experiments examine which training dynamics support transfer and how additional historical examples affect prediction.

\noindent\textbf{Obs. 5. The value of each training mechanism depends on the target dynamics.}
Figure~\ref{fig:tokenizer-ablation}(b) shows that removing recurrence causes the largest drop on Wiki, but barely affects ML-20M. Structural proximity matters most on ML-20M and YouTube, while preference and transitions have the largest effects on GoogleLocal and Flickr, respectively. Thus, \textbf{no single training mechanism provides the best preparation for every graph}. Although all variants retain the same tokenizer, removing a mechanism reduces transfer performance. This supports learning across different history--outcome relationships, rather than relying on a common representation alone.

\begin{figure}[h]
\centering
\includegraphics[width=\columnwidth]{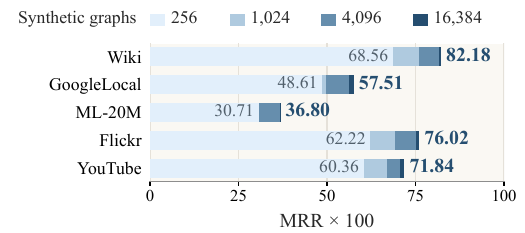}
\vspace{-15pt}
\caption{Training on more distinct synthetic graphs versus repeatedly using fewer graphs, with the same number of training updates and graph samples. Shades indicate graph-pool sizes; endpoints report MRR.}
\label{fig:prior-diversity}
\vspace{-10pt}
\end{figure}

\begin{figure}[t]
\centering
\includegraphics[width=\linewidth]{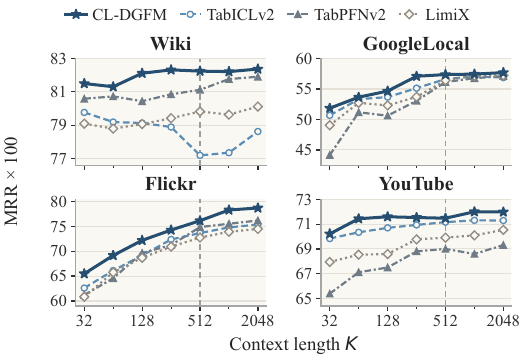}
\vspace{-20pt}
\caption{Effect of context size on prediction ($100\times$MRR).
$K$ counts positive context events, each paired with one negative.
Dashed lines mark the main setting $K=512$; vertical-axis ranges differ across datasets.}
\label{fig:context-efficiency}
\vspace{-10pt}
\end{figure}

\noindent\textbf{Obs. 6. Diverse synthetic graphs improve transfer beyond repeated training on a small graph pool.}
Figure~\ref{fig:prior-diversity} compares pools containing 256, 1,024, 4,096, and 16,384 distinct synthetic graphs. Each pool covers the same prior configurations, with variation in sampled graph structures and interaction dynamics. Every run uses 4,096 updates with four graphs per update, totaling 16,384 graph samples. The four settings therefore use each graph approximately 64, 16, 4, and 1 times, respectively. Expanding the pool improves MRR on all five real graphs, including from 68.56 to 82.18 on Wiki and from 62.22 to 76.02 on Flickr. These gains show that \textbf{learning from more realizations of the same dynamic mechanisms supports better transfer than repeatedly training on a small graph pool}. This supports generating varied interaction histories to prepare the shared predictor for different real networks.

\noindent\textbf{Obs. 7. Additional context helps some graphs substantially, but does not guarantee better prediction.}
Figure~\ref{fig:context-efficiency} shows strong context gains on Flickr and GoogleLocal, whereas Wiki and YouTube improve less and saturate earlier. On Wiki, frozen TabICLv2 even declines as context grows toward 512 positive events before partially recovering. Thus, \textbf{more labeled history does not automatically produce better interpretation of interaction patterns}. \method{} remains strongest across the tested sizes, with a clear advantage on Flickr but a narrowing gap on GoogleLocal as context grows. These results suggest that \textbf{learning across dynamics and supplying more target history offer different benefits}: context can reduce some transfer gaps, while the learned predictor remains important when additional examples yield limited gains.

\section{Conclusion}
We introduced \method{}, combining a common interaction language with learning across diverse synthetic dynamics. The language enables existing tabular predictors to process dynamic graphs, while labeled historical context supports prediction without target-graph parameter updates. Experiments on TGB and TGB-Seq demonstrate competitive performance across predictor families. These results support existing predictive architectures across dynamic networks through shared representations of interaction histories.

\newpage

\bibliographystyle{ACM-Reference-Format}
\bibliography{References-no-url-doi}

\appendix

\begin{table*}[h!]
\centering
\caption{Statistics of synthetic training data and real evaluation datasets. B/H denote bipartite/homogeneous graphs. Events counts all observed interactions; Test counts evaluation queries. Neg.\ denotes negative destinations per test query, excluding the positive destination.}
\label{tab:dataset-statistics}
\begingroup
\footnotesize
\setlength{\tabcolsep}{5pt}
\begin{tabular*}{\textwidth}{@{\extracolsep{\fill}}llrrrr@{}}
\toprule
Dataset & Domain (type) & Nodes & Events & Test & Neg. \\
\midrule
\multicolumn{6}{@{}l}{\textbf{Synthetic training data}} \\
Generated streams & Synthetic interactions (B/H) & $\approx 1.30$M & $\approx 28.50$M(training data) & -- & -- \\
\midrule
\multicolumn{6}{@{}l}{\textbf{TGB evaluation datasets}} \\
Wiki & Page edits (B) & 9,227 & 157,474 & 23,621 & 999 \\
Review & Product reviews (B) & 352,637 & 4,873,540 & 728,919 & 100 \\
Coin & Transactions (H) & 638,486 & 22,809,486 & 3,421,411 & 20 \\
Flight & Flights (H) & 18,143 & 67,169,570 & 10,026,943 & 20 \\
Comment & Social replies (H) & 994,790 & 44,314,507 & 6,647,176 & 20 \\
\midrule
\multicolumn{6}{@{}l}{\textbf{TGB-Seq evaluation datasets}} \\
GoogleLocal & Business reviews (B) & 473,580 & 1,913,967 & 43,546 & 100 \\
ML-20M & Movie ratings (B) & 110,431 & 14,494,325 & 129,291 & 100 \\
Flickr & Social links (H) & 233,836 & 7,223,559 & 609,679 & 100 \\
YouTube & Social links (H) & 402,422 & 3,288,028 & 232,865 & 100 \\
Yelp & Business reviews (B) & 1,743,769 & 19,760,293 & 1,032,354 & 100 \\
Taobao & Product clicks (B) & 1,623,633 & 18,853,792 & 2,406,071 & 100 \\
Patent$^\dagger$ & Citations (H) & 2,241,784 & 12,749,824 & 683,323 & 100 \\
WikiLink & Hyperlinks (H) & 1,361,972 & 34,163,774 & 3,073,045 & 100 \\
\bottomrule
\end{tabular*}
\par\smallskip
\begin{minipage}{\textwidth}
$^\dagger$Patent includes 1,931,005 history-only events in its event total; these events are not evaluation targets and are visible only to strictly later queries.
\end{minipage}
\endgroup
\end{table*}

\section{Datasets and Experimental Settings}
\label{app:experimental-protocol}

To facilitate reproducibility, we provide our code and experimental configurations in the following anonymous repository:
\url{https://anonymous.4open.science/r/CL-DGFM-C10D/}.

\subsection{Dataset Statistics and Splits}

Table~\ref{tab:dataset-statistics} reports dataset statistics, test-query counts, and negative-candidate counts. Node counts include only participating entities.
We use v2 of Wiki, Review, Coin, and Flight, and v1 of Comment. 
TGB~\cite{huang2023tgb} uses chronological splits with approximately 70\% for training, 15\% for validation, and 15\% for testing, whereas TGB-Seq~\cite{yi2025tgbseq} uses dataset-specific splits.
Together, these datasets evaluate whether a shared predictor transfers across interaction semantics, graph structures, and scales without target-graph parameter updates. They cover six application domains:

\begin{itemize}
\setlength{\itemsep}{0pt}
\setlength{\parsep}{0pt}

\item \textbf{Online interactions.}
Wiki records user--page edits, Comment captures social
replies, and Flickr and YouTube record social links.
These datasets cover user--content and user--user
interactions in bipartite and homogeneous graphs.

\item \textbf{Recommendation.}
Review, GoogleLocal, and Yelp record product or business reviews, ML-20M movie ratings, and Taobao product clicks. These bipartite graphs capture user preferences through feedback and clicks.

\item \textbf{Financial transactions.}
Coin records directed transactions between accounts. Each event identifies a source account, a destination account, and a timestamp, forming a temporal history of financial exchanges.

\item \textbf{Transportation.}
Flight records flights between airports, representing physical movements through a transport network. With approximately 67 million events, it contains the largest interaction history among our datasets.

\item \textbf{Citations.}
Patent records directed citations between patents. Its history also includes events that are not evaluation targets but remain observable to strictly later queries.

\item \textbf{Hyperlinks.}
WikiLink records links between Wikipedia pages, capturing changes in document connectivity. Its page-to-page relations complement the user--page editing activity recorded in Wiki.

\end{itemize}

\textbf{Our model is trained exclusively on synthetic interaction streams, and no real benchmark split is used to update its parameters. Evaluation uses the official test splits, with preceding real interactions providing observable history and labeled context without parameter updates.}

\subsection{Evaluation Details}

Observable history includes earlier validation and test events, but excludes interactions at or after the query timestamp. We rank the observed destination against the benchmark-provided negatives, retaining repeated candidate positions. MRR averages its reciprocal rank over test queries, with average ranks assigned to ties.

All target-graph contexts are sampled from the final 10\% of the training interval. 
Within each comparison, representations and predictors share context samples, labels, and test candidates. The context-size study varies only the context size, keeping each predictor's parameters fixed. 
Within each evaluation run, labeled context stays fixed, while observable history advances with query time. New interactions update subsequent candidate features without parameter updates or context reselection.

\noindent\textbf{Alternative representations.}
Representation comparisons use the same frozen TabICLv2 predictor. Topology contains 12 statistics: endpoint in/out neighborhood sizes, common neighbors, Adamic--Adar, resource allocation, Jaccard similarity, preferential attachment, forward/reverse pair counts, and three-hop path counts.

Time contains 20 statistics: the time since the latest event and event counts in four age intervals (0--0.03, 0.03--0.3, 0.3--3, and above 3, measured in calibrated time units) for each of the forward-pair, reverse-pair, source-incident, and destination-incident histories.
Topology+Time concatenates both groups. Appending candidate-relative ranks yields 24, 40, and 64 input dimensions, respectively.
Topology summarizes accumulated connectivity, whereas Time describes when interactions occurred.
Their concatenation tests whether combining conventional structural and temporal statistics suffices to match the tokenizer's explicit interaction patterns.

\noindent\textbf{Ablations.}
With model parameters fixed, tokenizer ablations separately remove memory, connectivity, or transition features, including the associated normalized features and matching candidate-relative rank columns, from both context and query inputs.
Prior ablations remove one generation mechanism at a time and retrain on the resulting synthetic streams, using the same tokenizer and matched training budgets.

\subsection{Baseline Sources and Implementation}

We reproduce CRAFT, TAMI, and TSNN using their public implementations and reported settings.
Other temporal baseline results are taken from the benchmarks.
In tokenizer-based predictor comparisons, \textbf{machine learning} and \textbf{deep learning} predictors take $\mathbf{X}^{ML\&DL}=\mathbf x$ as input, fit the labeled context, and select configurations by validation MRR.
For \textbf{foundation models}, we stack the same vectors $\mathbf x$ into an input table $X^{\mathrm{tab}}$, comprising labeled context rows and unlabeled query rows, and perform prediction with fixed parameters.
Main experiments run on NVIDIA A40 GPUs with Python 3.10, PyTorch 2.6.0, CUDA 12.4, and TabICL 2.2.0.

\section{Dynamic Interaction Tokenizer}
\label{app:tokenizer}

\subsection{Time Calibration and Feature Settings}

We calibrate time using the training history of each real
graph or the pre-context history of each synthetic task.
Let $n$ count events in this history, $n_{\mathrm{src}}$
count distinct sources, and $t_{\min},t_{\max}$ denote
its earliest and latest timestamps. The time scale is:
\begin{equation}
s=\max\left\{
\frac{(t_{\max}-t_{\min})n_{\mathrm{src}}}{n},
\epsilon_s
\right\}.
\label{eq:app-time-scale}
\end{equation}
We use $\epsilon_s=10^{-6}$ for real graphs and $10^{-8}$
for synthetic tasks, keeping $s$ fixed per graph or task.

All feature histories strictly precede the represented timestamp $t$. Let $C_{uv}$ count directed pair events and $d_x$ count all events incident to node $x$, including repeated interactions. We log-compress these counts using $\ell_+(x)=\log(1+x)$. $I_x$ retains the latest 256 incident timestamp entries of $x$, in either direction. 
$M_D(h)$ and $W_\ell(w,h)$ denote the uncompressed sums underlying Eqs.~\eqref{eq:decayed_memory} and~\eqref{eq:historical_connectivity_vector}.
Recency is $R_D=-\log(1+(t-\max D)/s)$, with $R_\varnothing=-30$. Empty sums are zero.
Table~\ref{tab:coordinates} follows the main-text concatenation order, with rows 9--18 forming
$\mathbf a(u,v,t)$. 
Rows 19--25 instantiate $\Omega_P$, and rows 26--30 instantiate $\Omega_T$.

\begin{table}[t]
\centering
\caption{The 31-dimensional history descriptor.
$\mathcal B=s\{0.03,0.3,3,30\}$ is ordered increasingly.
Ratios precede log compression, and $h=\infty$ removes decay.}
\label{tab:coordinates}
\begingroup
\scriptsize
\setlength{\tabcolsep}{2pt}
\renewcommand{\arraystretch}{1.1}
\begin{tabularx}{\columnwidth}{@{}l X@{}}
\toprule
No. & Features and settings \\
\midrule
\multicolumn{2}{@{}l}{
\textbf{Interaction memory: $\mathbf z^{\mathrm{mem}}\in\mathbb R^{18}$}
} \\
1--4 &
$\ell_+(M_{E_{uv}}(h)),\ h\in\mathcal B$,
using all past $u\to v$ events \\
5--8 &
$\ell_+(M_{I_v}(h)),\ h\in\mathcal B$ \\
9--11 &
$\ell_+(C_{uv}),\ \ell_+(d_u),\ \ell_+(d_v)$ \\
12--14 &
$R_{E_{uv}},\ R_{I_u},\ R_{I_v}$ \\
15--17 &
$\ell_+(C_{vu}),\
 \mathbf1\{C_{uv}>0\},\
 \mathbf1\{d_v=0\}$ \\
18 &
$\displaystyle
\ell_+\!\left(
\frac{M_{E_{uv}}(0.3s)}{1+M_{I_v}(0.3s)}
\right)$ \\
\midrule
\multicolumn{2}{@{}l}{
\textbf{Historical connectivity: $\mathbf z^{\mathrm{conn}}\in\mathbb R^7$}
} \\
19--21 &
$\ell_+(W_2(1,\infty)),\
 \ell_+(W_2(w_{\mathrm{act}},\infty)),\
 \ell_+(W_2(w_{\mathrm{act}},s))$ \\
22--24 &
$\ell_+(W_3(1,\infty)),\
 \ell_+(W_3(w_{\mathrm{act}},\infty)),\
 \ell_+(W_3(w_{\mathrm{act}},s))$ \\
25 &
$\ell_+(W_3(w_{\mathrm{dst}},\infty))$ \\
\multicolumn{2}{@{}p{\columnwidth}@{}}{
$w_{\mathrm{act}}(\rho)=
\prod_{k=1}^{\ell-1}\max(1,d_{x_k})^{-1}$
for intermediate nodes $x_k$, and
$w_{\mathrm{dst}}(\rho)=(1+d_v)^{-1/2}$.
} \\
\midrule
\multicolumn{2}{@{}l}{
\textbf{Ordered transitions: $\mathbf z^{\mathrm{seq}}\in\mathbb R^6$}
} \\
26--27 &
$\ell_+(T_1(\omega_{\mathrm{pos}},\infty)),\
 \ell_+(T_3(\omega_{\mathrm{pos}},\infty))$ \\
28 &
$\ell_+(T_3(\omega_{\mathrm{pos}},s))$ \\
29 &
$\ell_+(T_3(\omega_{\mathrm{src}},\infty))$ \\
30 &
$\ell_+(T_3(\omega_{\mathrm{dst}},\infty))$ \\
31 &
$\mathbf1\{v=v_u^{\mathrm{last}}\}$,
zero without outgoing history \\
\multicolumn{2}{@{}p{\columnwidth}@{}}{
$\omega_{\mathrm{pos}}=1/(ij)$,
$\omega_{\mathrm{src}}=
\omega_{\mathrm{pos}}/\sqrt{\max(1,d_b)}$,
and $\omega_{\mathrm{dst}}=
\omega_{\mathrm{pos}}/\sqrt{1+d_v}$,
where $i$ indexes references from most recent,
$j$ indexes eligible successors, and $b$ is their source.
} \\
\bottomrule
\end{tabularx}
\endgroup
\vspace{-10pt}
\end{table}

\subsection{History Retention}

To construct $\mathcal P_\ell(u,v,t)$, we expand paths
through each node's latest 256 incident entries and count
distinct event combinations separately.
The second hop excludes returns to $u$, and the third
excludes $u$ and the first intermediate node, with no
further deduplication.

To construct $\mathcal O_L(u,v,t)$, we inspect the latest 48 incident entries of each reference destination $a$ for starting visits $b\to a$ from other sources $b\ne u$.
For each starting visit at $\tau_a$, we scan 24 incident entries of $b$ from its first timestamp $\geq\tau_a$.
We retain the first four outgoing visits strictly between $\tau_a$ and $t$ to destinations other than $a$.
The successor index $j$ in $\omega(e)$ follows this retained order, regardless of which destinations appear in the query's candidate list.

\subsection{Input Transformation}

We use $c_x=4$ in Eq.~\eqref{eq:tokenizer_output}. Let $\operatorname{rank}_{q,k}(j)$ be candidate $j$'s ascending rank on feature $k$ within query $q$, starting at one and averaging ties. The rank vector in the main text is:
\begin{equation}
\mathbf r_{q,j}
=
\left[
\frac{2\operatorname{rank}_{q,k}(j)-1}{K_q}-1
\right]_{k=1}^{31}.
\label{eq:app-relative-rank}
\end{equation}
Repeated candidate positions are retained.
Context ranks are computed within each complete candidate
list before row selection.

\begin{algorithm}[t]
\caption{\method{}: Dynamic-Prior Learning and Cross-Graph Inference}
\label{alg:cl-dgfm}
\small
\begin{algorithmic}[1]
\Require Prior $\Pi$, pretrained backbone,
updates $U$, target graph $G$, target queries $\mathcal Q$
\Ensure Candidate scores $\{\mathbf s_q^G\}_{q\in\mathcal Q}$

\State Initialize the backbone and initialize the Contextual Value Adapter.
\State Freeze input encoders.

\Statex \textbf{Dynamic-prior learning}
\For{$k=1,\ldots,U$}
    \State Select generator configurations $\xi$
    according to $\Pi$.
    \State Generate sequences $\mathcal E$ using the
    source rates and destination distribution $\pi_\xi$
    in Section~\ref{sec:dynamic_prior}.
    \State Construct a batch of tasks $\mathcal T$ from
    earlier context events and strictly later queries.
    \ForAll{tasks $\mathcal T$}
        \State Construct $X_C,\mathbf y_C$ and
        $\{X_q\}_{q\in\mathcal I_Q}$ using the tokenizer.
        \State Apply one sampled, context-fitted
        preprocessing view.
        \State Score task queries with $F_\theta$
        using Eq.~\eqref{eq:contextual_candidate_scores}.
    \EndFor
    \State Estimate the batch ranking loss using
    Eq.~\eqref{eq:dynamic_training}.
    \State Update the in-context predictor and adapter
    with AdamW.
\EndFor

\Statex \textbf{Cross-graph inference}
\State Fix all model parameters.
\State Select historical context from $G$'s training
interval and construct $X_C,\mathbf y_C$.
\State Fit the backbone's two preprocessing views
on $X_C$.
\ForAll{$q\in\mathcal Q$}
    \State Construct $X_q$ for the given candidate list
    using the history visible before $q$.
    \State Run the adapted backbone under both views.
    \State Obtain $\mathbf s_q^G$ using prediction fusion
    and identical-descriptor averaging as specified in
    Appendix~\ref{app:contextual-modeling}.
\EndFor
\State \Return $\{\mathbf s_q^G\}_{q\in\mathcal Q}$.
\end{algorithmic}
\end{algorithm}

\section{Dynamic Interaction Prior}
\label{app:dynamic-prior}

\subsection{Event Generation}

For the source rates $\lambda_u(t)$, let $\mu_u>0$
be baseline rates with $\sum_u\mu_u=1$,
$\eta\geq0$ the excitation strength, and $\beta>0$
its decay rate. We use:
\begin{equation}
\lambda_u(t)
=
\mu_u+\eta\beta
\sum_{i:u_i=u,\ t_i<t}e^{-\beta(t-t_i)}.
\label{eq:app-source-intensity}
\end{equation}
Starting from empty histories, we generate event
times by thinning.

\subsection{Destination Mechanisms}

For $g_{\mathrm{pref}}$ and $g_{\mathrm{rec}}$,
let $c_u,c_v$ denote sampled community labels,
$B$ a positive community-affinity matrix,
and $a_v$ a positive destination propensity.
With two sampled decay scales $h_1,h_2>0$
and smoothing constant $\varepsilon=10^{-9}$,
their base scores are:
\begin{equation}
\begin{aligned}
g_{\mathrm{pref}}(u,v,t)
&=B_{c_u,c_v}a_v,\\
g_{\mathrm{rec}}(u,v,t)
&=\sum_{\tau\in E_{uv}}\sum_{k=1}^{2}
e^{-(t-\tau)/h_k}+\varepsilon.
\end{aligned}
\label{eq:app-preference-recurrence}
\end{equation}
Variants of $g_{\mathrm{pref}}$ incorporate destination-specific boosts, accumulated popularity, periodic modulation, or changes midway through the sequence.

For $g_{\mathrm{seq}}$ in Eq.~\eqref{eq:generative_mechanism_scores}, each row of $Q$ combines a weak background weighted by destination propensity with a few enhanced destinations, followed by normalization.
Without a previous destination, $g_{\mathrm{seq}}=g_{\mathrm{pref}}$.
For $g_{\mathrm{str}}$, we sample $\mathcal S_u(t)$ uniformly without replacement from eligible other sources and add $\varepsilon$ to the structural score.
We fall back to $g_{\mathrm{pref}}$ when closure is disabled or no eligible source or previous destination exists.

In Eq.~\eqref{eq:destination_prior}, normalization before mixing uses the full destination pool, and repetition inhibition is clipped to $\exp[-\min(\gamma C_{uv}(t),40)]$.
Self-links in homogeneous graphs are excluded before final normalization; bipartite graphs use a separate destination set.

\section{Contextual Modeling and Training}
\label{app:contextual-modeling}

\subsection{Tasks and Training}

We construct $\mathcal T$ from $\mathcal E$ using the
settings in Table~\ref{tab:context-training}, requiring
all queries to occur strictly after the context events.
Negative candidates exclude self-links and destinations
observed for the source at the represented timestamp.
Tasks alternate uniform negative sampling with a mixture
drawing up to half the negatives from previously visited
destinations. After tokenizing complete candidate lists,
we retain one positive and one negative per context event.

We train the in-context predictor and adapter with
Eq.~\eqref{eq:dynamic_training}, using positive-minus-negative
class logits as training scores and freezing input encoders.
One context-fitted backbone preprocessing view is sampled
per task.

\begin{table}[t]
\centering
\caption{Task construction and training settings.
Intervals refer to positions in the generated stream.}
\label{tab:context-training}
\begingroup
\footnotesize
\setlength{\tabcolsep}{3pt}
\renewcommand{\arraystretch}{1.08}
\begin{tabularx}{\columnwidth}{@{}l X@{}}
\toprule
Setting & Value \\
\midrule
Stream length &
$\{3{,}000,6{,}000,12{,}000\}$ events \\
Positive context events &
$\{64,128,256,512\}$ with probabilities
$(0.3,0.3,0.3,0.1)$ \\
Context / query interval &
$30$--$65\%$ / $65$--$95\%$ \\
Queries / candidates &
$8$ / $\{32,64,128\}$,
capped by the destination pool \\
Optimizer / learning rates &
AdamW, $10^{-5}$ for the predictor and
$10^{-3}$ for the adapter \\
Updates / tasks per update &
$1{,}024$ / $4$,
with $4{,}096$ updates for the diversity study \\
Random seeds & Three \\
Target context &
512 positive events from the final $10\%$ of the
training interval, each paired with one negative \\
\bottomrule
\end{tabularx}
\endgroup
\end{table}

\subsection{Contextual Value Adapter}

To specify $\operatorname{ValueAdapter}(\cdot)$ in
Eq.~\eqref{eq:contextual_value_modulation}, we summarize
the bounded memory-scale, connectivity, and transition
features in $X_C$ by their classwise means and standard
deviations under $\mathbf y_C$.
These statistics condition softmax-weighted combinations
of the four pair and four destination memory features.
The combined pair memories, their contrasts with
destination memories, and the connectivity and transition
features are standardized across context rows.
A learned affine map followed by $1+\tanh$ produces
a gate $\mathbf g_r\in(0,2)^{16}$ for each context row $r$.
Using $\bar{\mathbf h}_r$, the corresponding row
representation from $\bar H_C$, the projected correction is:
\begin{equation}
\begin{aligned}
\Delta\mathbf v_r
&=W_VW_{\mathrm{up}}
\left[(W_{\mathrm{down}}\bar{\mathbf h}_r)
\odot\mathbf g_r\right],\\
\Delta V_C
&=[\Delta\mathbf v_1,\ldots,
\Delta\mathbf v_{n_C}]^\top.
\end{aligned}
\label{eq:app-value-correction}
\end{equation}
Here, $W_{\mathrm{down}}$ and $W_{\mathrm{up}}$ project
into and out of a 16-dimensional subspace,
$\odot$ denotes elementwise multiplication, and $W_V$
is the backbone's original value projection.
This gives the correction added to $V_C$ in the main text.

\subsection{Target-Graph Inference}

Each target graph reuses its labeled context with fixed
model parameters.
The backbone averages aligned logits from two
context-fitted preprocessing views, \texttt{none}
and \texttt{power}, and applies binary softmax with
temperature $0.9$.
Before ranking, we average the resulting probabilities
among candidates with exactly identical 31-dimensional
descriptors within each query.
This assigns equal scores to tokenizer-indistinguishable
candidates without using labels or removing candidate
positions. Ties receive average ranks when computing MRR.

\section{Overall Algorithm}
\label{app:overall-algorithm}

Algorithm~\ref{alg:cl-dgfm} connects sequence generation,
task construction, and contextual prediction.
All input tables follow Eq.~\eqref{eq:tokenizer_output},
using only history strictly before each row's timestamp.
Candidate-relative ranks are computed before context-row
selection. The predictor $F_\theta$ includes the
Contextual Value Adapter.

\end{document}